\documentclass[conference]{IEEEtran}
\usepackage{cite}
\usepackage{amsmath,amssymb,amsfonts}
\usepackage{graphicx}
\usepackage{textcomp}
\usepackage{xcolor}
\def\BibTeX{{\rm B\kern-.05em{\sc i\kern-.025em b}\kern-.08em
    T\kern-.1667em\lower.7ex\hbox{E}\kern-.125emX}}

\usepackage{multirow}
\usepackage{caption}
\usepackage{subcaption}
\usepackage{colortbl}
\usepackage{graphicx}
\usepackage{enumitem}
\usepackage{multirow}
\usepackage{url}

\usepackage{algpseudocode}
\usepackage{xcolor}
\usepackage[linesnumbered,ruled,vlined]{algorithm2e}

\SetCommentSty{mycommfont}

\begin{document}

\title{Evolutionary Wave Function Collapse}


\author{
\IEEEauthorblockN{Dipika Rajesh}
\IEEEauthorblockA{
\textit{UC Santa Cruz}\\
Santa Cruz, USA\\
dipika.rajesh@gmail.com}
\and
\IEEEauthorblockN{Ahmed Khalifa}
\IEEEauthorblockA{
\textit{University of Malta}\\
Msida, Malta\\
ahmed@akhalifa.com}
\and
\IEEEauthorblockN{Julian Togelius}
\IEEEauthorblockA{
\textit{New York University}\\
New York, USA\\
julian@togelius.com}
}

\IEEEoverridecommandlockouts

\IEEEpubid{\makebox[\columnwidth]{979-8-3315-9476-3/26/\$31.00 \copyright2026 IEEE\hfill}
\hspace{\columnsep}\makebox[\columnwidth]{ }}

\maketitle

\IEEEpubidadjcol

\begin{abstract}
Wave Function Collapse (WFC) is a widely used procedural content generation method that learns local adjacency constraints from example inputs to generate larger outputs. In this paper, we explore combining WFC with evolutionary search by evolving the small input examples used by WFC rather than directly evolving complete levels. In this approach, WFC acts as a genotype-to-phenotype mapping. The generated levels are then evaluated through domain-specific fitness functions. We evaluate the method in two domains with different relationships between local and global structure: Maze connectivity maps and Zelda-style dungeon layouts. Our results show that evolutionary optimization over WFC inputs improves generation quality in domains where properties emerge from local relationships, while domains requiring global constraints remain challenging. These findings suggest that evolutionary search can effectively guide WFC generation when target objectives align with local structure.
\end{abstract}


\section{Introduction}
The wide range of Procedural Content Generation (PCG) methods in existence contains not only different algorithmic foundations, but also different use cases. For example, the Wave Function Collapse (WFC) algorithm ~\cite{karth2017wavefunctioncollapse}, which can be seen as constraint solving or self-supervised learning, is widely used in indie games. It is self-contained, requires little input, and rapidly generates small levels, although it cannot guarantee solvability and does not scale well to larger levels.

By contrast, search-based PCG sees content generation as an optimization to be solved with evolution, meaning that complex objectives and constraints can be satisfied. While search-based PCG is the subject of hundreds of papers, it is not widely used in actual games, probably because of the long generation time and/or complexity of specifying objectives.

In AI, as in other fields of engineering, the best methods are often hybrid methods that combine the advantages of multiple algorithms. What if we could combine some of the advantages of WFC and search-based PCG? For example, allowing authoring by specifying objectives. In this paper, we investigate whether the compact representation of WFC can be combined with the controllability of evolutionary search. Specifically, we evolve the inputs provided to WFC rather than directly evolving complete levels. This can be seen as using WFC as a genotype-to-phenotype mapping for the evolutionary algorithm. 

We evaluate this approach in two game domains with different relationships between local and global structure. In the Maze domain, where desired properties such as connectivity come out of local spatial relationships, evolutionary optimization successfully utilizes WFC toward higher-quality outputs. In the Zelda domain, where the game depends on global constraints such as exactly one player, key, and door, the limitations of local patterns become more apparent. Our results suggest that evolutionary optimization over WFC genotypes is most effective when target objectives are compatible with local structural regularities.

\section{Background}
Procedural content generation~\cite{shaker2016procedural} is getting a lot of attention from the game industry in the past couple of years\footnote{\url{https://steamdb.info/stats/releases/?tagid=1716}}. The problem is that most of the techniques that are used in the industry are based on human knowledge and heuristics, which makes the algorithm not generalizable between games, as each game requires a different algorithm. Wave Function Collapse (WFC) is one of the early methods that can be treated as a black box and used in your game with high controllability of the generated output through using a small example picture.

\subsection{Wave Function Collapse}
Wave Function Collapse (WFC) was introduced by Maxim Gumin in 2016\footnote{\url{https://github.com/mxgmn/WaveFunctionCollapse}}. WFC has gained plenty of attention since, from both game developers and academia. WFC is a constraint-solving method~\cite{karth2017wavefunctioncollapse} that tries to generate bigger patterns based on the input constraints. These constraints can be provided directly through a tile mode or can be learned from a smaller image input in the pixel mode. The pixel mode can be considered as self-supervised learning~\cite{summerville2018procedural} as the algorithm learns the constraints from the input image.  

WFC uses a sliding window to learn local pattern relations without any understanding of global patterns. For example, WFC can't learn that a certain tile needs to be only available a fixed number of times or that there is a path between two tiles on the map. This simplicity of the input/output of WFC made it at the forefront of PCG methods used in commercial games~\cite{stalberg2018wfc} and ROM hacking communities~\cite{mawhorter2021content}. 

Several extensions to WFC have explored methods for improving controllability, scalability, and output quality. Prior work has incorporated global constraints~\cite{sandhu2019enhancing,cheng2020automatic}, constructive layout generation~\cite{minini2020combining}, neural representations~\cite{karth2021neurosymbolic}, hierarchical generation~\cite{alaka2023hierarchical}, graph-based variants of WFC~\cite{kim2019automatic}, and modified tile selection heuristics to better preserve input tile distributions~\cite{bateni2023better}. Other work has combined WFC with reinforcement learning or evolutionary approaches to improve generated outputs~\cite{babin2021leveraging,bailly2022genetic}. These works highlight a broader interest in combining WFC with optimization and controllability methods.

\subsection{Search-Based Procedural Content Generation}
Search-based procedural content generation (SBPCG)~\cite{togelius2011search} formulates the generation process as an optimization problem. Evolutionary algorithms and related search methods optimize generated artifacts according to game related metrics such as playability, connectivity, progression structure, or difficulty. Search-based approaches have been applied to levels, rulesets, and music generation~\cite{shaker2013ropossum,togelius2008experiment,scirea2016metacompose}.

Compared to constructive approaches such as WFC, search-based methods provide significantly more control over generated content because designers can explicitly optimize toward desired objectives. However, these approaches are often computationally expensive and may require carefully designed fitness functions. Prior work has also explored searching for generators rather than directly searching for levels~\cite{kerssemakers2012procedural,khalifa2020multi}. Our work follows this direction by evolving compact WFC input examples that act as generators for larger procedurally generated outputs.

\section{Methods}
Our method combines Wave Function Collapse (WFC) with evolutionary search by treating WFC inputs as evolvable generators. We directly evolve the small input image (each pixel represent a game tile) that are used by WFC to generate larger outputs. In this method, each input acts as a genotype, while the generated WFC level acts as the phenotype evaluated by the fitness function. WFC in this case serves as a stochastic genotype-to-phenotype mapping within an evolutionary algorithm.

\subsection{Representation}
Each individual in the population is represented as a small tile grid. In all the experiments, we use a $4 \times 4$ genotype representation initialized with random tiles from the target game. The genotype is rendered as an image and passed to WFC, which extracts adjacency patterns and generates a larger output level. This representation allows evolutionary search to optimize on compact examples instead of directly optimizing complete levels that may be larger and more computationally expensive.

\subsection{Wave Function Collapse Generation}
For each genotype, WFC generates an output level by extracting all the 2x2 patterns in the input image and use them as constraints for the generation. Generation is stochastic, meaning that the same genotype can produce different outputs depending on the random seed. In our experiments, each genotype is evaluated using a single WFC generation. We evaluate two game domains:
\begin{itemize}
    \item \textbf{Maze}: generates $8 \times 8$ maze-like maps focused on connectivity and navigable path structure.
    \item \textbf{Zelda}: generates $16 \times 16$ dungeon-like layouts containing players, keys, doors, and enemies.
\end{itemize}

\subsection{Fitness Evaluation}
Generated levels are evaluated using domain-specific fitness functions from the PCG Benchmark~\cite{khalifa2025procedural}. For the Maze domain, fitness rewards levels that contain a single connected traversable region and long navigable paths. For the Zelda domain, fitness is a weighted sum of an entity-count term and a playability term: $\text{fitness} = 0.7 \cdot R_{\text{entity}} + 0.3 \cdot R_{\text{play}}$. The $R_{\text{entity}}$ term rewards count close to the target (one player, one key, one door, up to 25 enemies). $R_{\text{play}}$ is zero unless exactly one player, key, and door are present. Otherwise, it averages reachability between player, key, and door, plus a bonus when the combined path length is near half the level area.

\subsection{Evolutionary Search}
We use an evolutionary algorithm with tournament selection, mutation, and elitism. Each run uses a population size of 10 over 100 generations. At each generation:
\begin{enumerate}
    \item Each genotype is evaluated by generating a WFC output and computing its fitness.
    \item Tournament selection (size 3) selects $N/2 = 5$ parents.
    \item Each selected parent contributes two individuals to the next generation: an unmutated copy and a mutated offspring, maintaining a constant population size of $N = 10$.
    \item For each mutated offspring, between 1 and 5 tile positions are independently mutated, depending on the population's fitness standard deviation. Each candidate mutation is applied with probability $p$, and replaces the selected tile with a value drawn uniformly at random from the tileset. The mutation probability $p$ starts at 0.5 and adapts each generation based on whether the fitness has improved to balance exploration and exploitation.
    \item The best individual of the generation is preserved into the next generation (elitism).
\end{enumerate}

\subsection{Random Search Baseline}
We compare the evolutionary algorithm against a random search baseline. In random search, each generation consists of a newly sampled random population and keep the best chromosomes. Both methods use the same evaluation of 100 generations with a population size of 10, resulting in 1000 WFC genotype evaluations per run.

\section{Results}
\begin{figure*}[t]
    \centering
    \begin{subfigure}{0.2\textwidth}
        \centering
        \includegraphics[width=\linewidth]{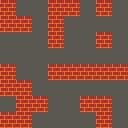}
        \caption{Maze EA}
    \end{subfigure}
    \hfill
    \begin{subfigure}{0.2\textwidth}
        \centering
        \includegraphics[width=\linewidth]{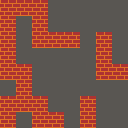}
        \caption{Maze Random}
    \end{subfigure}
    \hfill
    \begin{subfigure}{0.2\textwidth}
        \centering
        \includegraphics[width=\linewidth]{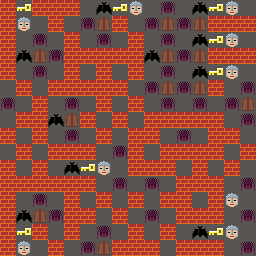}
        \caption{Zelda EA}
    \end{subfigure}
    \hfill
    \begin{subfigure}{0.2\textwidth}
        \centering
        \includegraphics[width=\linewidth]{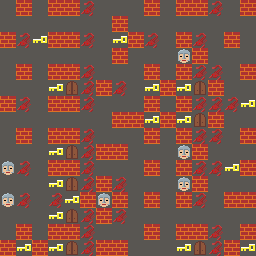}
        \caption{Zelda Random}
    \end{subfigure}

    \caption{
    Outputs generated using evolutionary and random search in the Maze and Zelda domains respectively. Evolutionary search produces more coherent and connected layouts in the Maze domain compared to random search. In the Zelda domain, evolutionary search generates more structured arrangements of the entities and traversable regions, although satisfying global gameplay constraints remains challenging due to WFC’s reliance on local relationships.
    }

    \label{fig:results}
\end{figure*}

\begin{figure}[t]
    \centering
    \includegraphics[width=0.9\linewidth]{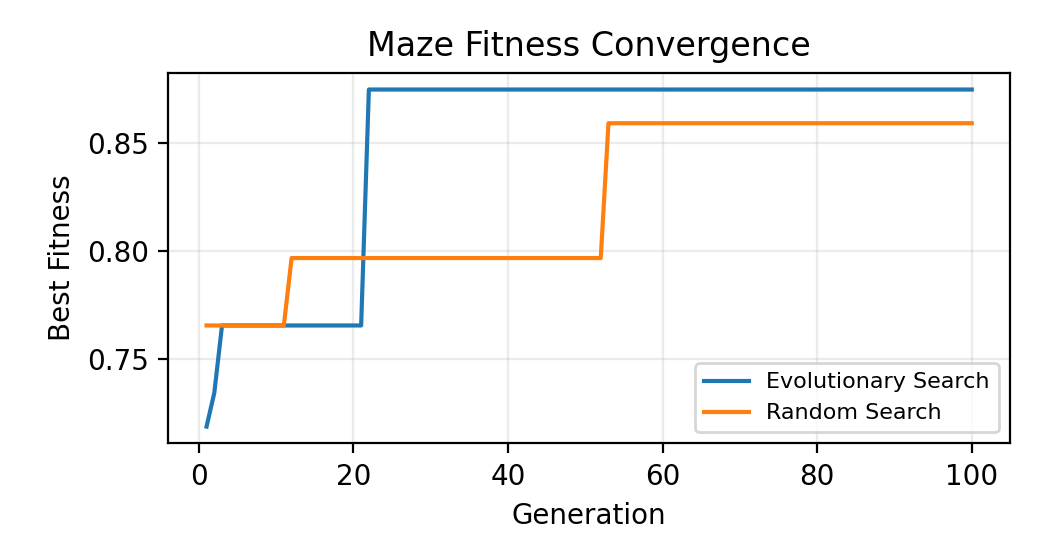}
    \caption{
    Best fitness over generations in the Maze domain. Evolutionary search rapidly converges toward high-performing WFC inputs, while random search only improves as it samples better inputs.
    }
    \label{fig:maze_convergence}
\end{figure}

Figure~\ref{fig:results} shows the outputs generated using evolutionary and random search in both the Maze and Zelda domains. The two domains showcase different relationships between local and global structure. This allows us to evaluate how effectively evolutionary search can guide WFC generation.

\subsection{Maze Domain}
In the Maze domain, evolutionary search consistently produced more coherent and connected layouts than random search. Evolved outputs typically contained longer continuous paths and fewer disconnected regions, while random search frequently generated fragmented layouts with large and isolated components.

Figure~\ref{fig:maze_convergence} shows the fitness progression in the Maze domain. Evolutionary search rapidly converges toward high-performing generators within the first 40 generations before stabilizing near a local optimum. In contrast, random search exhibits large variance throughout the generations and fails to showcase consistent improvement over time.

\begin{figure}[t]
    \centering
    \includegraphics[width=0.9\linewidth]{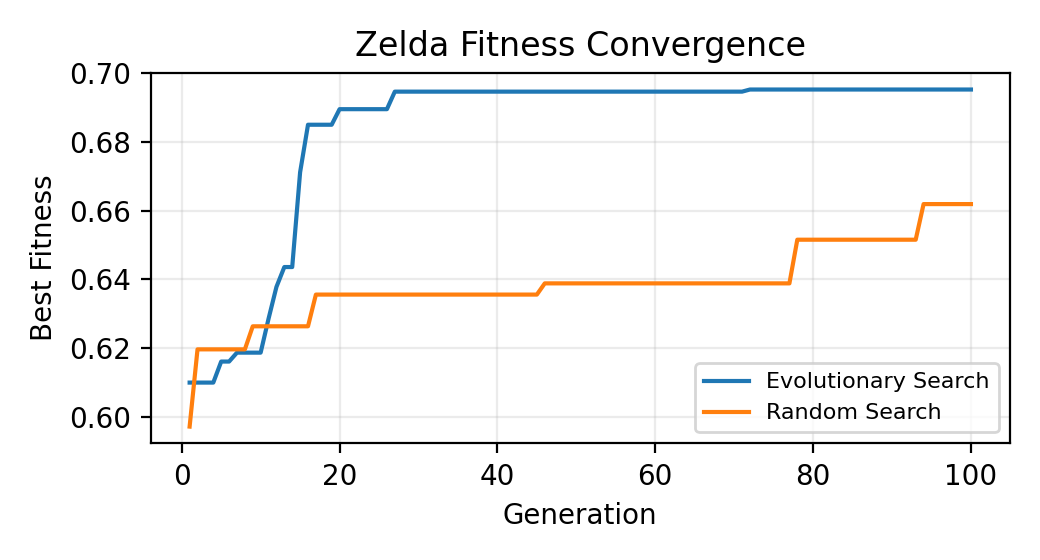}
    \caption{
    Best fitness over generations in the Zelda domain. Evolutionary search rapidly improves during the early generations before converging near a local optimum, while random search advances in sparse jumps and converges more slowly towards a lower plateau.
    }
    \label{fig:zelda_convergence}
\end{figure}

Although random search occasionally discovers strong outputs through random luck or chance, these generators are not typically preserved between generations. In contrast, evolutionary search retains and mutates successful WFC inputs, allowing local spatial patterns to compound over time. The Maze domain appears particularly compatible with the local pattern representation used by WFC. Desired gameplay properties such as connectivity emerge naturally from repeated local adjacency relationships. These results suggest that evolving WFC inputs behave similarly to searching for local construction rules. The evolutionary algorithm discovers compact examples that allow WFC to tend toward generating desired larger structures.

\subsection{Zelda Domain}

The Zelda domain proved significantly more difficult. Although evolutionary search generated more structured outputs than random search, many generated levels still failed to satisfy all gameplay requirements at the same time. The evolved outputs frequently showed recognizable room-like regions, traversable corridors, and more meaningful placement of enemies and gameplay entities, and wall and maze-like structures. However, making sure of the existence of exactly one player, one key, and one door remained challenging. 

Figure~\ref{fig:zelda_convergence} shows the fitness progression in the Zelda domain. Evolutionary search improves rapidly during the early generations before stabilizing near generation 30. In contrast, random search slowly improves or even stagnates throughout the run because the best successful generators are not utilized overall towards improvement.

Quantitatively, evolutionary search also achieved higher fitness values than random search in the Zelda domain. However, the performance gap is smaller than in the Maze domain. This highlights the difficulty of optimizing on gameplay properties that depend on global structure. This limitation follows from the local nature of WFC. The algorithm learns adjacency relationships between nearby patterns but does not explicitly model global gameplay dependencies. Consequently, evolutionary search is constrained by the representational limits of the underlying WFC model.

Despite these limitations, evolutionary search still improved the overall organization of generated Zelda layouts relative to random search. The evolved examples appeared to shift WFC toward producing more playable spatial arrangements, even when complete gameplay validity was not achieved.

\subsection{Discussion}
The contrast between the Maze and Zelda domains highlights an important property of Evolutionary Wave Function Collapse. The approach performs best when the objectives follow local spatial regularities. In such domains, evolution can successfully search for compact WFC inputs that expand into coherent outputs reliably. Domains requiring stronger global constraints remain more challenging. Even in situations where evolution discovers useful local patterns, WFC itself cannot directly enforce global requirements such as uniqueness constraints or progression ordering.

Another important limitation is the stochastic nature of WFC generation. Because each genotype can generate different outputs depending on the random seed, fitness evaluation introduces noise into the evolutionary process. In this work, each genotype was evaluated using a single WFC generation. Evaluating each genotype multiple times may improve stability and reduce variance in future work.

\section{Conclusion \& Future Work}

This paper explored combining Wave Function Collapse with evolutionary search by evolving compact WFC input examples rather than directly evolving complete levels. Our results show that evolutionary optimization can effectively guide WFC generation in domains where gameplay properties emerge from local structure. In the Maze domain, evolution successfully discovered compact WFC inputs that consistently generated connected and navigable layouts. In the Zelda domain, evolutionary search improved the organization of gameplay entities and traversable structure, but satisfying global gameplay constraints remained significantly more challenging due to the local nature of WFC pattern learning.

These results suggest that Evolutionary Wave Function Collapse is particularly effective in domains where global properties can emerge indirectly from repeated local relationships. However, domains that depend on explicit and specific symbolic constraints expose important limitations of purely local pattern models.

Future work should explore integrating explicit global constraint solving~\cite{sandhu2019enhancing,cheng2020automatic} into the evolutionary WFC pipeline. Instead of relying entirely on local adjacency relationships, future systems could incorporate constraint-matching methods that directly enforce gameplay requirements such as uniqueness constraints, progression ordering, reachability, or key-door dependencies during or after generation.

\bibliographystyle{IEEEtran}
\bibliography{references}

\end{document}